\pdfoutput=1
\documentclass[sigconf]{acmart}

\usepackage{comment}
\usepackage{multirow} 
\usepackage{makecell}
\usepackage{stfloats}
\usepackage{enumitem}

\begin{document}
\title{Learning Generalizable Latent Representations for Novel Degradations in Super Resolution}







\author{Fengjun Li$^{2,*}$, Xin Feng$^{2,*}$, Fanglin Chen$^{2}$, Guangming Lu$^{1,2}$ and Wenjie Pei$^{2, \dag}$}
\thanks{* Both authors contributed equally to this research.}
\thanks{$\dag$ Corresponding author.}
\affiliation{
  \institution{$^1${Guangdong Provincial Key Laboratory of Novel Security Intelligence Technologies}}
  \institution{$^2${Harbin Institute of Technology} \city{Shenzhen} \country{China}}
}
\email{20s151173@stu.hit.edu.cn,{fengx_hit,wenjiecoder}@outlook.com, {chenfanglin, luguangm}@hit.edu.cn}

\renewcommand{\shortauthors}{Fengjun Li and Xin Feng, et al.}

\begin{abstract}
Typical methods for blind image super-resolution (SR) focus on dealing with unknown degradations by directly estimating them or learning the degradation representations in a latent space. A potential limitation of these methods is that they assume the unknown degradations can be simulated by the integration of various handcrafted degradations (e.g., bicubic downsampling), which is not necessarily true. The real-world degradations can be beyond the simulation scope by the handcrafted degradations, which are referred to as novel degradations. In this work, we propose to learn a latent representation space for degradations, which can be generalized from handcrafted (base) degradations to novel degradations. The obtained representations for a novel degradation in this latent space are then leveraged to generate degraded images consistent with the novel degradation to compose paired training data for SR model. Furthermore, we perform variational inference to match the posterior of degradations in latent representation space with a prior distribution (e.g., Gaussian distribution). Consequently, we are able to sample more high-quality representations for a novel degradation to augment the training data for SR model. We conduct extensive experiments on both synthetic and real-world datasets to validate the effectiveness and advantages of our method for blind super-resolution with novel degradations.
\end{abstract}




\keywords{Super resolution, degradation, latent representation}

\maketitle
\section{Introduction}

\begin{figure}[!t]
  \centering   
  \includegraphics[width=0.98\linewidth]{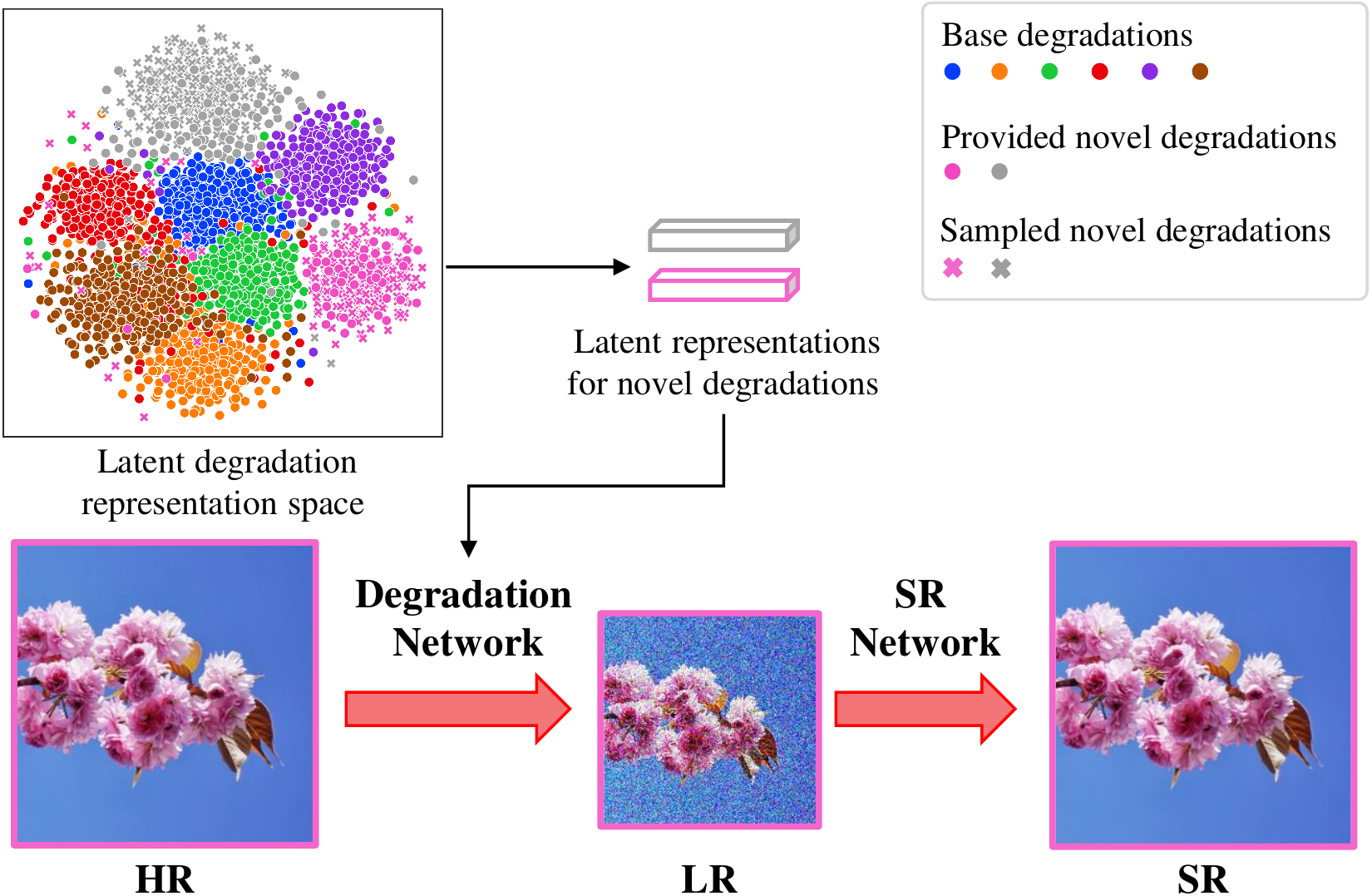}
  \vspace{-10pt} 
   \caption{We learn a latent degradation representation space which can generalize from handcrafted base degradations with paired training data to novel degradations beyond the simulation scope of base degradations. We further perform variance inference to match the posterior of degradations in the latent space with a Gaussian distribution, which allows sampling more high-quality degradations for novel degradations. The learned degradation representations are further used to guide the HR-LR degradation for generating paired training samples for learning SR network.}
   \label{fig:intro}
   \vspace{-14pt} 
\end{figure}

Blind image super-resolution (SR) aims to predict the high-resolution image from a low-resolution image with unknown degradations, such as blur, noise, JPEG compression, etc. It is a fundamental research topic in Computer Vision and has extensive applications ranging from medical imaging~\cite{App-medical}, security~\cite{App-security} to astronomical images~\cite{App-astronomical}.

Despite the rapid progress in image super-resolution based on deep learning, blind image SR remains an extremely challenging task due to the unpredictability and diversity of degradations. 

Most existing methods for blind image SR seek to either estimate the unknown degradations directly~\cite{KMSR,RealSR_KM} or predict degradation representations in a latent space~\cite{IKC,DAN}. These methods are designed based on the assumption that the unknown degradations can be simulated by the integration of various handcrafted degradations (e.g., bicubic downsampling). A prominent example is DASR~\cite{DASR_cl}, which applies contrastive learning to learn a latent representation space for degradations. By sampling sufficiently diverse handcrafted degradations to compose training data for contrastive learning, DASR performs well on the unknown degradations that lie within the simulation scope by the enumerated handcrafted degradations during training. However, the real-world degradations can be beyond such simulation scope, in which case DASR can hardly predict the representations precisely for such novel degradations. 

Another way to handle novel degradations that is beyond the simulation scope by handcrafted degradations, is to employ Generative Adversarial Networks (GANs)~\cite{GAN} to learn the novel degradations directly~\cite{DegradationGAN,FSSR,DASR}, and then generate the degraded images consistent with the novel degradations for composing paired training data for downstream SR model. However, the performance of such method is heavily limited by the generation error of GANs and meanwhile, it can only handle one degradation with one discriminator. 

In this paper we propose to learn a latent representation space for degradations, which can be generalized from handcrafted base degradations to novel degradations that are beyond the simulation scope of base degradations. To this end, we design three pretext tasks to guide the learning of latent representation space for degradations. First, we perform classification on base degradations to make different degradations be distinguishable in this latent space. Second, we perform unsupervised categorization on novel degradations via adversarial learning to generalize the latent space from base degradations to novel degradations. Finally, we perform variational inference to match the posterior distribution of degradations with a prior distribution (e.g., Gaussian distribution), which allows sampling more high-quality representations for novel degradations to augment training data for learning downstream SR model. 

The obtained representations for a novel degradation in this latent space are then leveraged to generate degraded images consistent with the novel degradation to compose paired training data for learning the downstream SR model. 
To conclude, we make following contributions. 1) The Generalizable Degradation Representation Learner is proposed to learn a generalizable latent representation space from base degradations to novel degradations that are beyond the simulation scope by the base degradations.
2) The Degradation-Consistent HR-LR-SR Generative Network is designed to leverage the learned representations for a novel degradation to generate paired HR-LR training data for learning SR model in the phase of LR-SR super-resolution. 3) Extensive experiments validate the effectiveness of our methods both quantitatively and qualitatively. 

\vspace{-4px}
\section{related work}

\noindent\textbf{SR with known degradations.} Since the pioneering work for CNN-based single image SR is introduced by Dong \textit{et al.} \cite{SRCNN}, extensive works have been proposed to improve SR performance of LR images with fixed bicubic degradation. These methods aim to explore more effective techniques to learn the mapping from LR to HR image, which include but not limited to such aspects: network architecture\cite{VDSR,FSRCNN,DRCN,LapSRN}, objective function \cite{PerceptualLoss,EnhanceNet,SRGAN,ESRGAN,CinCGAN} and training strategy\cite{ESRGAN,ProSR,SRFBN}. 
Beyond the single-degradation SR, many works attempt to address known multiple degradations. 
Inspired by Efrat \textit{et al.}~\cite{Efrat2013AccurateBM} which suggests that accurate estimation of degradations is more essential than sophisticated image priors for single image SR, many existing works aim to estimate the degradations. SRMD~\cite{SRMD} incorporates degradation map of blur kernel and noise level as an extra input to perform SR. Then Luo \textit{et al.}~\cite{UDVD} introduce dynamic convolutions and propose a refine network called UDVD, producing better results than SRMD. Gu \textit{et al.}~\cite{IKC} proposes IKC to correct degradation estimation. Recently, Wang \textit{et al.}~ \cite{DASR_cl} propose DASR, which extracts degradation representation from LR images using contrastive learning, and the representation is used to predict convolutional kernels for SR.

\begin{figure*}[!t]
    \centering
    \vspace{-6pt}
    \includegraphics[width=0.75\linewidth]{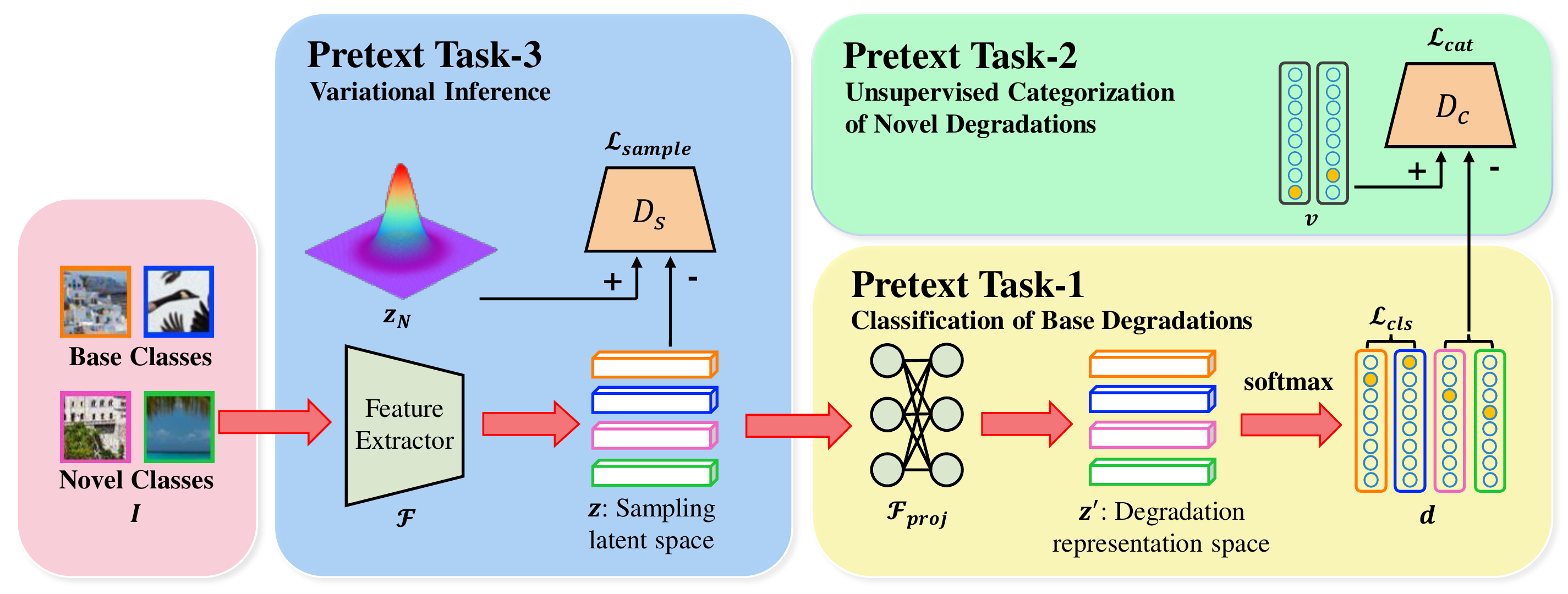}
    \vspace{-12px}
    \caption{Framework of Generalizable Degradation Representation Learner. Three pretext tasks are designed to guide the learning of both sampling latent space and the degradation representation space.}
    \label{fig:stage1}
    \vspace{-12px}
\end{figure*}

\noindent\textbf{Blind SR with unknown multiple degradation.} Although current works \cite{SRMD, UDVD,IKC,DASR_cl} have shown improved performance when the multiple degradation model is predetermined, there is still a distinct gap between synthesized LR and real-world LR images. One way to address this problem is to collect real-world paired data with different parameters \cite{Camera_SR, Real_SR} for training. However, these methods cannot cover all kinds of scenarios. Another category of works investigates unsupervised image SR. One branch of approaches \cite{KMSR,RealSR_KM} assumes that each of real-world degradations can be modeled by specific kernel, thus they seek to build a large kernel pool via kernel estimation to generate realistic LR from HR images for paired training. However, these methods cannot generalize to arbitrary degradations such as compression due to strong degradation assumption. Another branch of approaches proposes to learn the degradation using Generative Adversarial Networks (GANs)~\cite{GAN}. Yuan \textit{et al.}~\cite{CinCGAN} first introduce such idea and propose CinCGAN to learn the distribution of clean LR and HR image separately, which guides the SR of unpaired LR images later. Nonetheless, it only employs cycle-consistency loss for maintaining image content in LR space, leading to limited performance for SR. Unlike Yuan \textit{et al.}~\cite{CinCGAN}, several works~\cite{DegradationGAN,pseudo_SR,FSSR,DASR} propose to learn the degradations to generate LR from HR images firstly, then the obtained HR-LR paired data are used to train the SR network in a supervised manner. However, the performance of such method is heavily limited by the generation error of GANs and meanwhile, it can only handle one degradation with one discriminator. 

In this work, we propose to learn a latent space for degradation representation, which can be generalized from base degradations to novel degradations that cannot be simulated by the base degradations. The learned latent representation for a novel degradation is then leveraged to generate paired HR-LR training pair for learning downstream SR model. 

\vspace{-4px}
\section{Method}

\begin{figure*}[!t]
    \centering
    \vspace{-3pt}
    \begin{minipage}[b]{0.9\linewidth}
    \includegraphics[width=\linewidth]{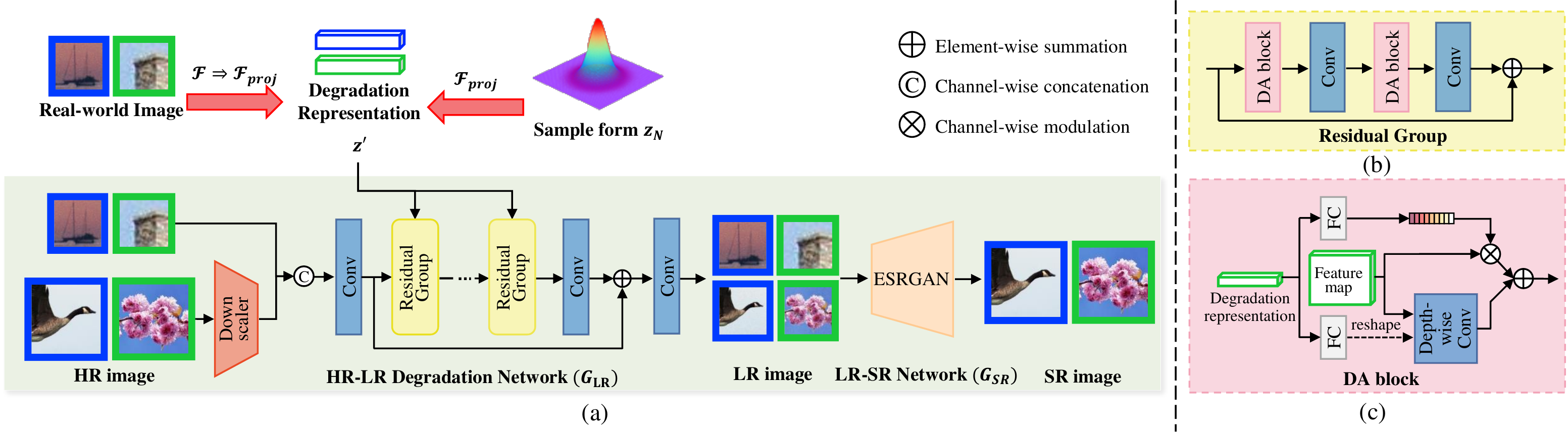}
    \end{minipage}
    \vspace{-12pt}
    \caption{Architecture of HR-LR-SR generative networks. The learned degradation representations by our \emph{GDRL} are incorporated into the HR-LR degradation to generate LR images for composing paired training data for downstream LR-SR super-resolution.}
    \label{fig:stage2}
    \vspace{-12pt}
\end{figure*}

We aim to learn a latent representation space for potential degradations, which can be generalized from handcrafted base degradations to novel degradations that are beyond the simulation scope by the base degradations. The learned representation for a novel degradation in this latent space is leveraged to generate degraded low-resolution (LR) images consistent with the novel degradation from high-resolution (HR) images to compose paired training data. Then the obtained training data is utilized to learn the downstream super-resolution (SR) model. To this end, Our method consists of two modules: 1) Generalizable Degradation Representation Learner and 2) Degradation-Consistent HR-LR-SR Generative Network, which are illustrated in Figure~\ref{fig:stage1} and Figure~\ref{fig:stage2} respectively.


\vspace{-4pt}
\subsection{Generalizable Degradation Representation Learner}
\vspace{-2pt}
We design the Generalizable Degradation Representation Learner (\emph{GDRL}) to learn a latent representation space for 
encoding degradations with three goals. First, the latent space should be distinguishable between different degradations. Second, considering the gap between the handcrafted degradations and the real-world degradations, the learned latent space for degradations should have well generalizability from known base degradations to novel degradations. Third, we aim to perform sampling in this latent space for a novel degradation to obtain more representation samples for augmenting the training data for downstream HR-LR-SR generative network. To achieve these goals, we design three pretext tasks correspondingly to learn latent representation space for degradations.  

Formally, given a degraded LR image $I$, the proposed \emph{GDRL} first encodes it by projecting it into the constructed latent feature space by a feature extractor:
\vspace{-3pt}
\begin{equation}
\vspace{-3pt}
    \mathbf{z} = \mathcal{F}(I),
    \label{eqn:project}
\end{equation}
where $\mathcal{F}$ denotes the feature extractor, which comprises the convolutional layers of ResNet-18~\cite{ResNet} and a fully-connected layer. $\mathbf{z}$ is the encoded representation for $I$.

\noindent\textbf{Pretext task-1: classification of handcrafted (base) degradations.}
To optimize the latent space to make representations for different degradations separable between each other, we synthesize degraded images with various handcrafted base degradations and then perform degradation classification on the encoded representations of the degraded images. Thus, the degradation classification can be viewed as a pretext task to perform supervised learning on the latent space parameterized by $\mathcal{F}$. Given a 
latent representation $\mathbf{z}$ for $I$ obtained by Equation~\ref{eqn:project}, we employ Cross Entropy loss (CE) to optimize the learning of $\mathcal{F}$:
\vspace{-4pt}
\begin{equation}
\vspace{-4pt}
\begin{split}
    & \mathbf{z}^\prime = \mathcal{F}_{\text{proj}}(\mathbf{z}), \\
    &\mathcal{L}_\text{cls} = \text{CE}(y_I, \mathcal{F}_{\text{softmax}}(\mathbf{z}^\prime)),
\end{split}
\label{eqn:representation}
\end{equation}
where $y_I$ is the degradation label for $I$ and $\mathcal{F}_{\text{softmax}}$ denotes the Softmax function. Note that we can obtain $y_I$ since $I$ is generated with a base degradation.  $\mathcal{F}_{\text{proj}}$ is a lightweight feature project module consisting of three fully connected layers and the LeakyReLU~\cite{LeakyRelu} in between. Because the latent space constructed by $\mathcal{F}$ is used for sampling from a prior distribution which is performed by Pretext task-3, we project the latent representation $\mathbf{z}$ from the sampling latent space $\mathcal{F}$ to a new feature space by $\mathcal{F}_\text{proj}$, called the degradation representation space, for classification to loose the coupling between $\mathcal{F}$ and the classification task. 

\noindent\textbf{Pretext task-2: unsupervised categorization of novel degradations via adversarial learning.}
While the base degradations can be encoded in the learned latent space for degradation representation based on pretext task 1, the novel degradations, which are beyond the simulation scope by base degradations, cannot be properly encoded. To generalize the latent space from base degradations to novel degradations, we perform unsupervised categorization in the latent space on the degraded images with novel degradations that are to be super-resolved.

Since the novel degradations to be super-resolved are unknown, we can only perform categorization in an unsupervised way. Specifically, we employ an adversarial network to impose a categorical distribution and thereby encourage the degraded samples with the same novel degradation to be clustered together. Given a degraded image $I$ with a novel degradation and its latent representation $\mathbf{z}^\prime$ obtained by Equation~\ref{eqn:representation}, we employ Softmax function to obtain a vectorial category embedding $\mathbf{d} \in \mathbb{R}^K$:
\vspace{-2pt}
\begin{equation}
\vspace{-2pt}
    \mathbf{d} = \mathcal{F}_{\text{softmax}}(\mathbf{z}^\prime).
    \label{eqn:category}
\end{equation}
Then we train an adversarial network to guide the category embedding $\mathbf{d}$ to be close to a one-hot encoding:
\vspace{-2pt}
\begin{equation}
\vspace{-2pt}
\small
\begin{split}
    \mathcal{L}_\text{cat} =& \min_{\emph{\text{GDRL}}}\max_{\mathcal{D}_c}\mathbb{E}_{\mathbf{v}\sim p_\text{one-hot}}[\log \mathcal{D}_c(\mathbf{v})] \\   &+\mathbb{E}_{\mathbf{d}\sim p_{\mathbf{d}}}[\log (1-\mathcal{D}_c(\mathbf{d}))],
\end{split}
\end{equation}
where $\mathcal{D}_c$ denotes the discriminator in the adversarial network and $\mathbf{v} \in \mathbb{R}^K$ is a one-hot encoding drawn from the one-hot encoding distribution.
Note that only the discriminator in the adversarial network needs to be trained since our Generalizable Degradation Representation Learner (\emph{GDRL}) serves as the generator. 

It should be noted that the dimension $K$ of $\mathbf{d}$ is a hyper-parameter, which is equal to the sum of the base degradation categories used in pretext task-1 and reserved category number for the novel degradations. We cannot precisely predict the exact category number of the novel degradations. Typically larger $K$ implies more reserved category number for the novel degradations and leads to more fine-grained clustering. 
In our implementation, we explicitly categorize the samples with novel degradations into new categories other than the base degradations. 
In the case that a novel degradation happens to be similar to one of the base degradations, it can be considered as a fine-grained categorization within this category.


\noindent\textbf{Pretext task-3: variational inference via adversarial learning.}
Given limited degraded images for a novel degradation, we aim to not only learn the latent representation for this degradation, but also sample more high-quality samples in the latent space to obtain more training data with this novel degradation for the downstream HR-LR-SR generative process. To this end, we perform variational inference to push the posterior distribution of all samples to follow a prior distribution. Then we can perform sampling to achieve more samples with the novel degradations.

We adopt the similar way of performing variational inference as adversarial autoencoder (AAE)~\cite{AAE}, which uses adversarial training to impose a prior distribution on the posterior distribution in the latent space. To be specific, we learn a discriminator $\mathcal{D}_s$ by providing the samples $\mathbf{z}_N$ drawn from a Gaussian distribution in the sampling latent space by $\mathcal{F}$ as positive samples and the encoded samples $\mathbf{z}$ by our \emph{GDRL} as negative samples. The discriminator $\mathcal{D}_s$ and our \emph{GDRL} are trained in an adversarial manner:
\begin{equation}
\begin{split}
    \mathcal{L}_\text{sample} = &\min_{\emph{\text{GDRL}}}\max_{\mathcal{D}_s}\mathbb{E}_{\mathbf{z}_N\sim \mathcal{N}(0,1)}[\log \mathcal{D}_s(\mathbf{z}_N)] \\
    &+ \mathbb{E}_{\mathbf{z}\sim p_{\mathbf{z}}}[\log (1-\mathcal{D}_s(\mathbf{z}))].
\end{split}
\end{equation}
The sampled latent representations are further fed into the downstream HR-LR degradation process to achieve more HR-LR paired training data for LR-SR learning process.

\begin{figure}[!t]
    \centering
    \begin{minipage}[b]{1\linewidth}
    \includegraphics[width=1.0\linewidth]{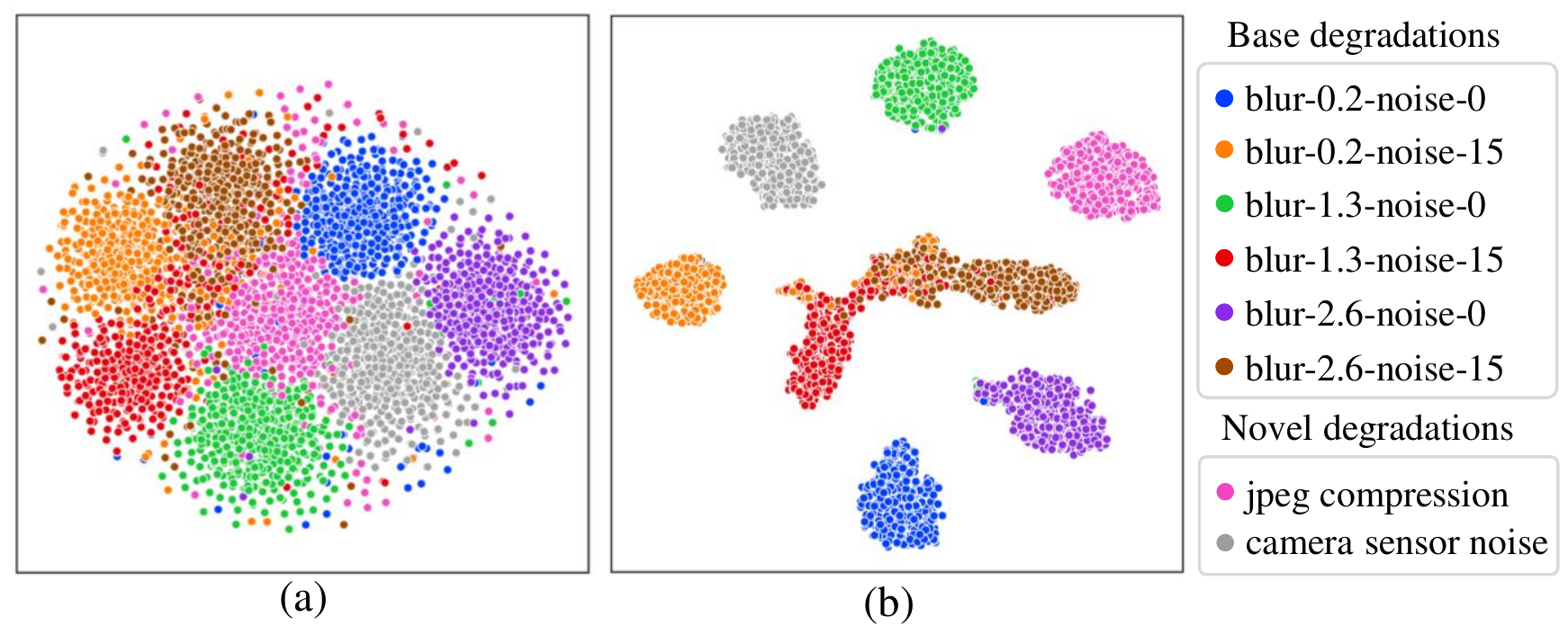}
    \end{minipage}
    \vspace{-14pt}
    \caption{t-SNE maps of sample distributions in the sampling space constructed by $\mathcal{F}$ and the degradation representation space projected by $\mathcal{}{F}_\text{proj}$ from $\mathcal{F}$, respectively. Samples of 6 base degradations and 2 novel degradations are visualized.}
    \label{fig:sampling_class}
    \vspace{-14pt}
\end{figure}

Note that we perform variational inference and sample in the sampling latent space constructed by $\mathcal{F}$ whilst using the latent space projected by $\mathcal{F}_\text{proj}$ from $\mathcal{F}$ as the final degradation representation space. 
This is mainly because the degradation representation space is more distinguishable than the sampling space for different degradations. Figure~\ref{fig:sampling_class} presents the t-SNE visualization of same set of samples in the sampling space and the degradation representation space, respectively.

All three pretext tasks can be performed jointly to supervise our \emph{GDRL} in an end-to-end manner:
\vspace{-3pt}
\begin{equation}
\vspace{-2pt}
    \mathcal{L}_\text{GDRL} = \mathcal{L}_\text{sample} + 
    \alpha \mathcal{L}_\text{cls} + \beta \mathcal{L}_\text{cat} 
\end{equation}
where $\alpha$ and $\beta$ are hyper-parameters to balance three losses.

\vspace{-4pt}
\subsection{Degradation-Consistent HR-LR-SR Generative Network}
\vspace{-2pt}
For the novel degradations that we aim to super-resolve, there is no paired HR-LR images for training. Thus we first perform HR-LR degradation from arbitrary HR images, which is guided by the learned representations of novel degradations with our \emph{GDRL}, to generate degraded images consistent with the novel degradations. As a result, we are able to compose paired HR-LR training data for learning the SR model during the LR-SR super-resolution. Such HR-LR-SR generative framework is similar to Bulat et al.~\cite{DegradationGAN}. 

\subsubsection{HR-LR Degradation}
Following DASR~\cite{DASR_cl}, we incorporate the latent degradation representations into the generative process by predicting the convolutional kernels from degradations. We adopt the similar network structure as DASR in our HR-LR degradation module except that we first down-sample the HR image to the scale of LR image by bicubic operation. As shown in Figure~\ref{fig:stage2}, the HR-LR degradation module iteratively stacks 5 residual groups, each of which consists of 5 DA blocks designed in DASR. In each DA block, the degradation representation is used to predict the kernel of depth-wise convolutions and channel-wise coefficients.

To optimize the HR-LR degradation module, we conduct two individual training modes in parallel. In the first mode, we feed a HR image along with the latent representation of a specified novel degradation into the HR-LR degradation module and apply supervision to guide the generated LR image to follow the specified degradation. Meanwhile, the content of the LR image should be consistent with the input HR image. In the second mode, a LR image with the specified novel degradation is fed into the HR-LR degradation module without downsampling, the HR-LR degradation module is supervised to reconstruct the input LR image. Thus, the second training mode acts like an autoencoder, which guides the HR-LR degradation module to recognize and reserve the degradation-sensitive features during the encoding and decoding process. In both training modes, we employ two types of loss functions to ensure the degradation consistency and content consistency for generated LR images.


\begin{figure*}[!t]
    \centering
    \vspace{-3pt}
    \begin{minipage}[b]{1\linewidth}
    \centering
    \includegraphics[width=1\linewidth]{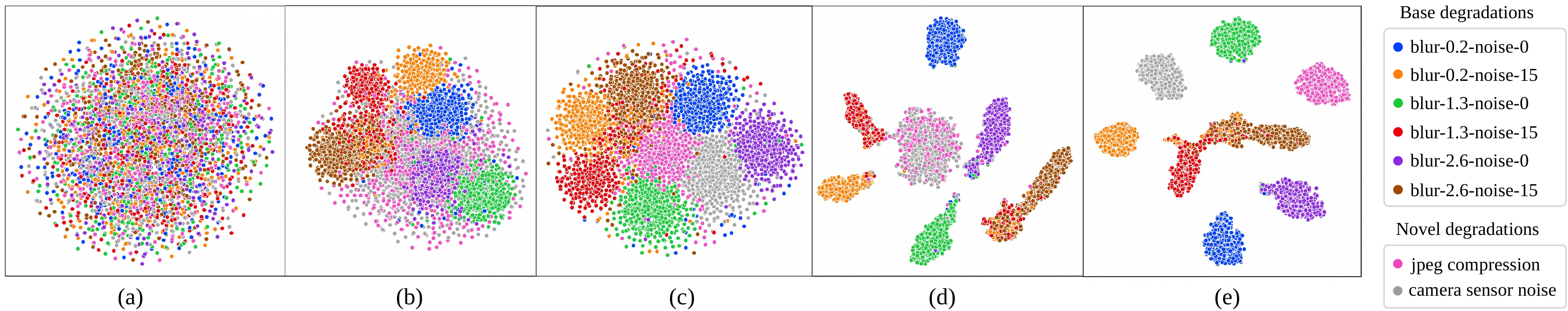}
    \end{minipage}
    \vspace{-18pt}
    \caption{t-SNE maps of sample distributions in the latent sampling space optimized using (a) only pretext task-3 (variational inference to guide the samples to follow a Gaussian distribution), (b) both pretext task-3 and task-1 (classification of base degradations), and (c) all three pretext tasks, respectively. (d) and (e) visualize the t-SNE distributions of samples in the degradation representation space optimized without and with pretext task-2, respectively, while both pretext task-1 and task-3 are used in these two cases.}
    \label{fig:ablation1}
    \vspace{-12pt}
\end{figure*}

\noindent\textbf{Degradation consistency.} To make the generated LR image be degraded following the specified degradation, we design three loss functions. First, we obtain the vectorial category embedding $\mathbf{d} \in \mathbb{R}^K$ for a generated LR image $I_{l}$ using our trained \emph{GDRL} as shown in Equation~\ref{eqn:category}, which can be viewed as the pseudo degradation label. Then we employ  cross-entropy loss (CE) between them and the specified groundtruth degradation $y_{I}$:
\begin{equation}
    \mathcal{L}_\text{rep} =  \text{CE}\Big(\mathcal{F}_\text{softmax}(\mathcal{F}_\text{proj}(\mathcal{F}(I_{l}))), y_{I_l}\Big),
\end{equation}

where $\mathcal{F}_\text{proj}$ and $\mathcal{F}$ are the transformation functions in \emph{GDRL} to obtain the degradation representation from an input degraded image. Second, we perform adversarial supervision along with an auxiliary classifier using the similar way as AC-GAN~\cite{ACGAN}. Specifically, we learn a discriminator $\mathcal{D}_\text{rep}$ to distinguish between the high-frequency of generated LR images and that of real degraded images and thereby push the HR-LR degradation network (denoted as $G_\text{LR}$) to generate consistent degraded images with the real degraded images:
\begin{equation}
\begin{split}
    \mathcal{L}_\text{GAN-LR} = &\min_{G_\text{LR}}\max_{\mathcal{D}_\text{rep}}\mathbb{E}_{I_\text{real}\sim p_\text{real}}[\log \mathcal{D}_\text{rep}(h(I_\text{real)})] \\
    &+ \mathbb{E}_{I_l \sim p_{G_\text{LR}}}[\log (1-\mathcal{D}_\text{rep}(h(I_l)))].
\end{split}
\end{equation}
Herein, $h(x)$ is a high-pass filter following FSSR~\cite{FSSR}.

Meanwhile, an auxiliary classifier $\mathcal{F}_\text{ac}$ for degradation categories, which consists of 3 ResBlock\cite{ResNet} and 2 fully-connected layers, is trained along with the discriminator $D_\text{rep}$ in an adversary manner using the cross-entroy loss (CE):
\begin{equation}
        \mathcal{L}_\text{ac} = \text{CE}( \mathcal{F}_{\text{softmax}}(\mathcal{F}_\text{ac}(I)), y_I).
\end{equation}

\noindent\textbf{Content consistency.} Three loss functions are used to ensure HR-LR content consistency. For the first training mode that degrades a HR image into a LR image, we apply the Color loss~\cite{FSSR} and the perceptual loss~\cite{PerceptualLoss}. Color loss performs L1 constraints between a generated LR $I_l$ and the degraded image $I_l^\prime$ from the HR image by bicubic downsampling:
\begin{equation}
    \mathcal{L}_\text{color} = \| I_l - I_l^\prime\|_1.
\end{equation}

Perceptual loss~\cite{PerceptualLoss} ($\mathcal{L}_\text{per}$)is used to minimize the difference at the semantic level in the deep feature space of VGG-19\cite{VggNet}.

For the second training mode which acts as an autoencoder, we employ pixel-wise L1 reconstruction loss to minimize the difference between the generated LR image and the input groundtruth LR.



\subsubsection{LR-SR Super-Resolution}
Using the degraded LR images by the HR-LR degradation module, we are able to compose paired HR-LR training data for learning the downstream LR-SR super-resolution module. As a result, any SR model that is trained with paired training data can be used as our LR-SR super-resolution module. In our implementation, we opt for ESRGAN\cite{ESRGAN} which is a classical super-resolution model trained with paired data.  
We train the LR-SR super-resolution module following the training of ESRGAN\cite{ESRGAN}, using three losses: 1) $L_1$ reconstruction loss to minimize the pixel-wise $L_1$ distance between the generated SR and HR groundtruth, 2) perceptual loss~\cite{PerceptualLoss} to reduce their semantic distance and 3) adversarial loss to push the generated SR to be as realistic as the paired HR by learning a discriminator $\mathcal{D}_\text{SR}$:
\begin{equation}
\begin{split}
    \mathcal{L}_\text{GAN-SR} = &\min_{G_\text{SR}}\max_{\mathcal{D}_\text{SR}}\mathbb{E}_{I_\text{HR}\sim p_\text{HR}}[\log \mathcal{D}_\text{SR}(I_\text{HR})] \\
    &+ \mathbb{E}_{I_s \sim p_{G_\text{SR}}}[\log (1-\mathcal{D}_\text{SR}(I_s))],
\end{split}
\end{equation}
where $G_\text{SR}$ denotes the LR-SR super-resolution module. Three losses are balanced with the weights tuned on a validation set.

\vspace{-4px}
\section{Experiments}
\begin{figure*}[!t]
    \centering
    \begin{minipage}[b]{0.80\linewidth}
    \includegraphics[width=1\linewidth]{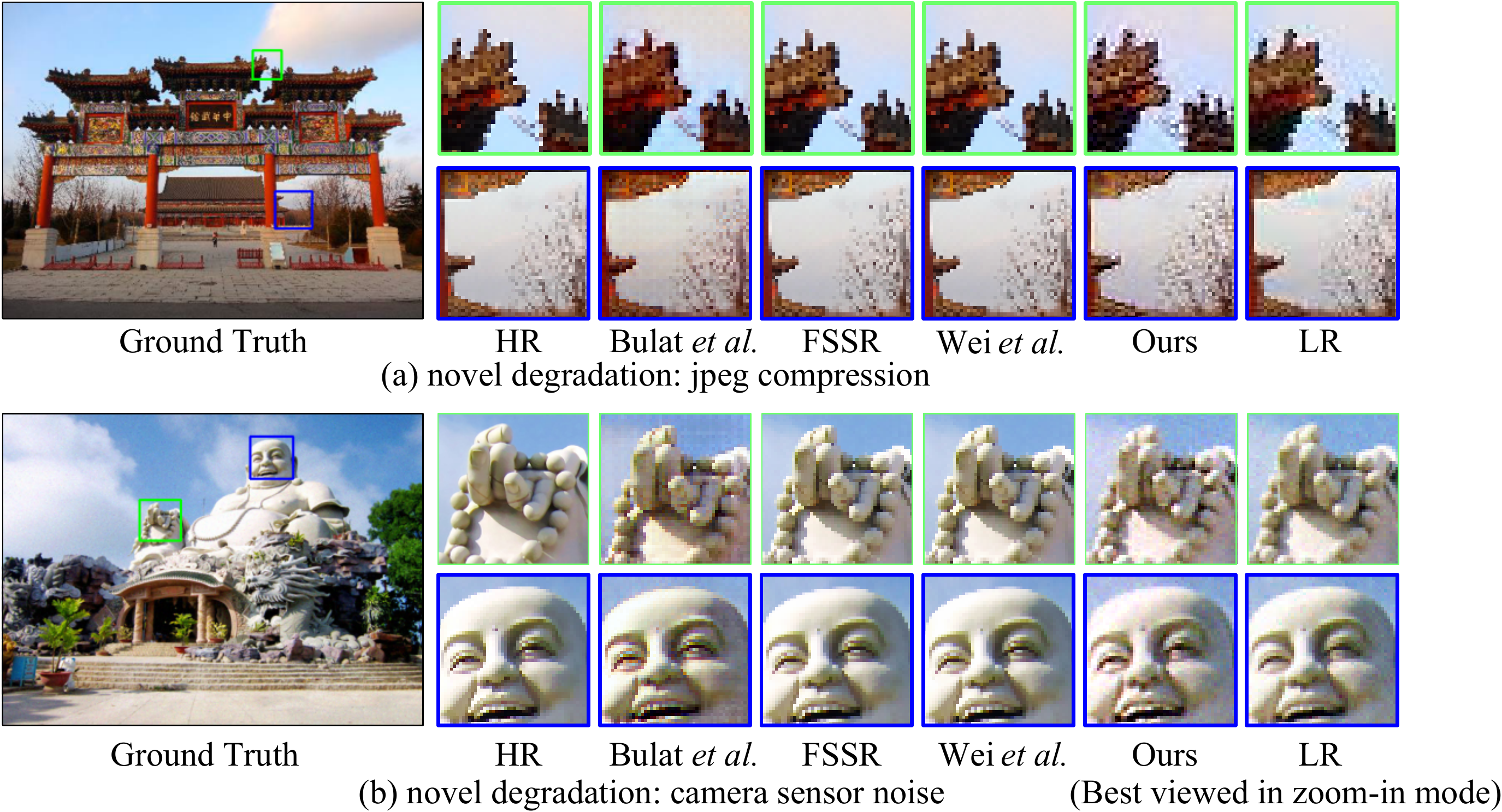}
    \end{minipage}
    \vspace{-10pt}
    \caption{Degraded LR images (4$\times$) by our method and other 3 baselines for two handcrafted novel degradations, respectively.}
    \vspace{-10pt}
    \label{fig:result1_sy}
\end{figure*}

\subsection{Datasets and Implementation Details}

\noindent\textbf{Base degradations.} Following the typical way of handcrafting the degradations~\cite{UDVD}, we consider 6 typical handcrafted degradations as base degradations by integrating different degrees of Gaussian blurry and different levels of Gaussian noise. Concretely, we set 3 levels of Gaussian blurry with fixed size of Gaussian kernel $15 \times 15$: \{0.2, 1.3, 2.6\} in terms of the Gaussian variance. Meanwhile, we set two levels of Gaussian noise: \{0, 15\} in terms of Gaussian variance. Then we can obtain 6 different degradations by combinations of different levels of Gaussian blurry and Gaussian noise. We use `blur-0.2-noise-15' to denote the degradation with Gaussian blurry of 0.2 variance and Gaussian noise of 15 variance.

\noindent\textbf{Novel degradations.} To ensure that the novel degradations are beyond the simulation scope of the base degradations, we use two sets of novel degradations: 1) handcrafted novel degradations and 2) real-world degradations. Both sets contain two degradations that differs substantially from the base degradations. To be specific, in the set of handcraft degradations, the first novel degradation is called `jpeg compression', which is handcrafted by 4x downsampling and then jpeg compression with a quality of 30. the second novel handcrafted degradation is `camera sensor noise'~\cite{BSRGAN}, which is obtained by applying reverse-forward ISP processing along with 4x downsampling. In the set of real-world degradations, we use two classical datasets: `AIM2019 dataset'~\cite{AIM2019} and `NTIRE2020 dataset'~\cite{NTIRE2020}, each of which contains one type of unpaired degraded images with essentially different degradation from the based degradations.

We use 800 images from DIV2K~\cite{DIV2K} and 2,650 images from Flick2K~\cite{EDSR} as the original HR images for obtaining the LR images for base degradations and handcrafted novel degradations, respectively. 
During training, LR images are cropped into patches of size 64×64 with augmentation by randomly flipping and rotation. The size of HR patches is 256×256 corresponding to scale factor 4. Adam\cite{Adam} is used for optimization. Code reproducing the results of our experiments will be available online.

\vspace{-6pt}
\subsection{Ablation Study}

\noindent\textbf{Effect of Pretext task-1 (classification of base degradations).}
We first perform ablation study to investigate the effect of pretext task-1, which is classification of base degradations for optimizing the latent space to be separable between different base degradations.
Figure~\ref{fig:ablation1} visualizes the t-SNE distribution maps in the sampling latent space (by $\mathcal{F}$ in Equation~\ref{eqn:project}) for samples with 8 different degradations (6 base and 2 novel). Figure~\ref{fig:ablation1} (a) and (b) are the distributions when the sampling latent space is optimized with only pretext task-3 and with both pretext task-1 and pretext task-3, respectively. It reveals that the pretext task-1 enables the learned sampling space to distinguish between different base degradations.

\begin{figure}[t]
    \centering
    \begin{minipage}[b]{1.0\linewidth}
    \includegraphics[width=1.0\linewidth]{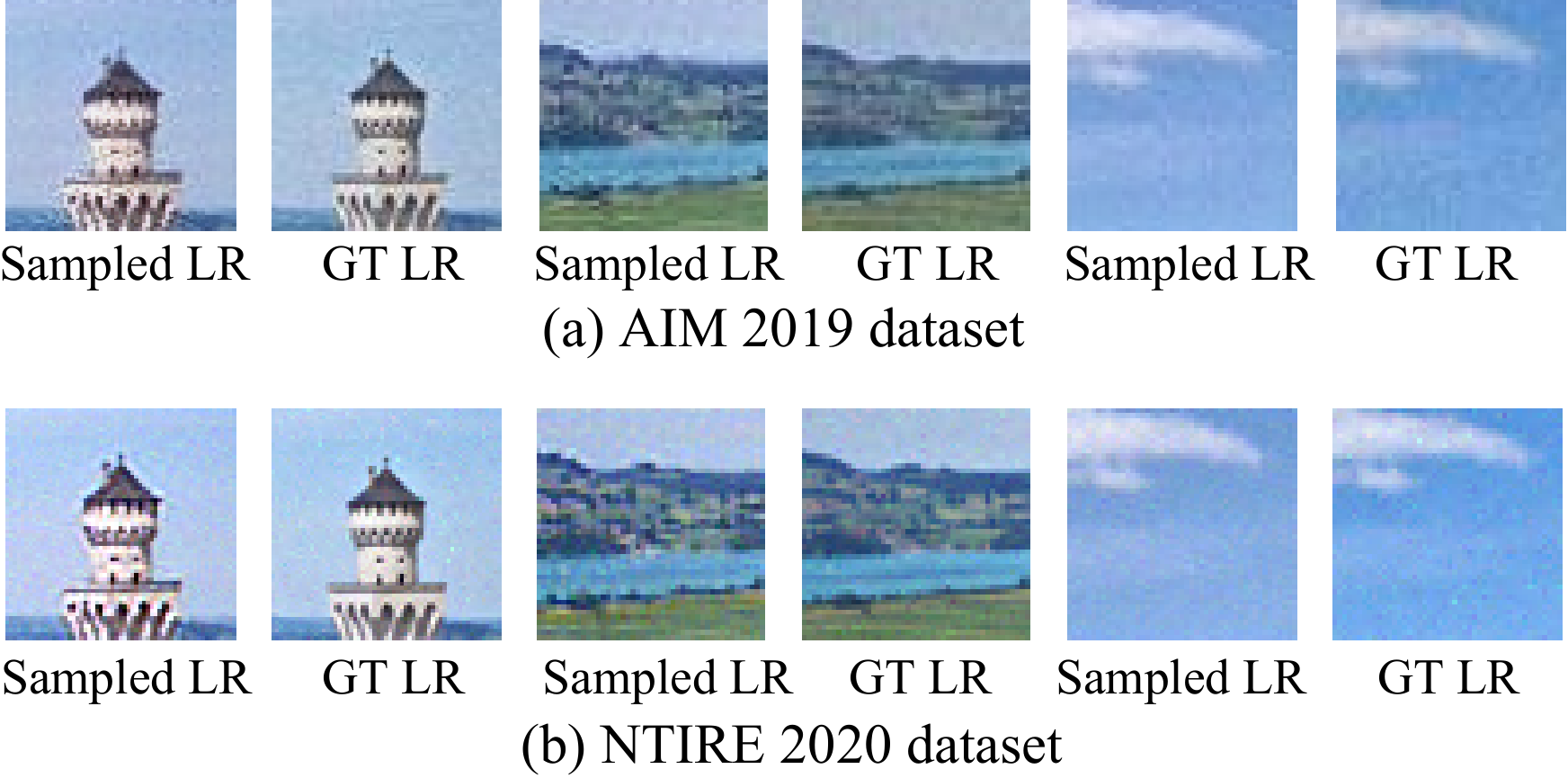}
    \end{minipage}
    \vspace{-18pt}
    \caption{Degraded LR images (4$\times$) with the sampled degradation representations for both two real-world datasets. The groundtruth (GT) real LR images are provided for reference.}
    \label{fig:sample}
    \vspace{-8pt}
\end{figure}

\noindent\textbf{Effect of Pretext task-2 (unsupervised categorization of novel degradations).}
Figure~\ref{fig:ablation1} (c) shows the t-SNE visualization of sample distribution in the sampling latent space optimized using all three pretext tasks. Comparison between the distributions in Figure~\ref{fig:ablation1} (b) and (c) manifests that the pretext task-2 enables our model to cluster the samples for each novel degradation together. We also visualize the distribution of samples in the degradation representation space optimized without and with the pretext task-2 respectively in (d) and (e) of Figure~\ref{fig:ablation1}, where samples with 2 novel degradations cannot be separated from each other. Besides, some samples with novel degradations are falsely clustered into the set of base degradations, which can be addressed using the pretext task-2.

\noindent\textbf{Effect of Gaussian sampling by pretext task-3.}
We perform variational inference with pretext task-3 to match the posterior distribution in the sampling space to Gaussian distribution. Thus we can sample more degradation representations for a novel degradation and obtain more paired training data for SR model. Figure~\ref{fig:sample} presents several degraded LR images with sampled degradation representations for both real-world novel degradations, the groundtruth (real) LR images are provided for reference. We observe that the degraded LR images with the sampled degradations have the consistent degradation patterns with the groundtruth LR images. Furthermore, table~\ref{tab:SR} shows the comparative results of SR performance of our method with or without the sampled novel representations for training. It shows that using sampled novel representations boosts the performance moderately in terms of all three metrics for both handcrafted and real-world degradations. 



\begin{figure*}[!t]
    \begin{minipage}[b]{0.81\linewidth}
    \includegraphics[width=1\linewidth]{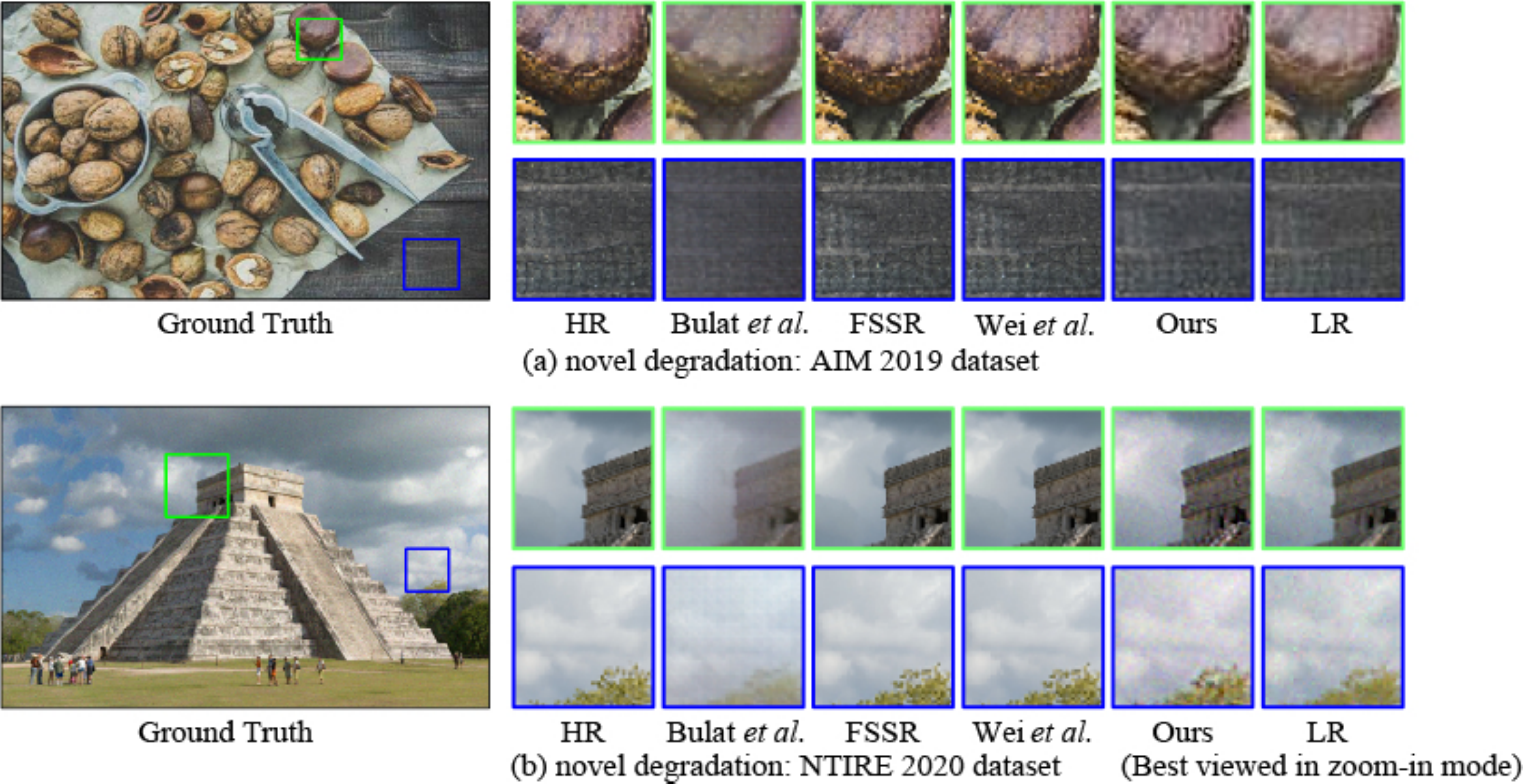}
    \end{minipage}
    \vspace{-6pt}
    \caption{Degraded LR images (4$\times$) by our method and other 3 methods for two real-world degradations, respectively.}
    \vspace{-8pt}
    \label{fig:result1_real}
\end{figure*}

\begin{table*}[t]
\caption{Evaluation of degraded LR (4$\times$) by our method and other 3 baselines for both handcrafted and real-world degradations.}
\vspace{-8pt}
\centering
\renewcommand\arraystretch{1.0}
\resizebox{0.88\linewidth}{!}{
    \begin{tabular}{l|c|ccc|ccc|ccc|ccc}
        \toprule
        \multirow{3}{*}{Method} & 
        \multirow{3}{*}{Scale} &
        \multicolumn{6}{c|}{Handcrafted degradations}&
        \multicolumn{6}{c}{Real-world degradations}\\
        & &\multicolumn{3}{c|}{Camera sensor noise}  & 
        \multicolumn{3}{c|}{Jpeg compression} &
        \multicolumn{3}{c|}{AIM 2019 dataset}  & 
        \multicolumn{3}{c}{NTIRE 2020 dataset}\\
          & & PSNR$\uparrow$ & SSIM$\uparrow$ & LPIPS$\downarrow$  & PSNR$\uparrow$        & SSIM$\uparrow$      & LPIPS$\downarrow$ & PSNR$\uparrow$        & SSIM$\uparrow$      & LPIPS$\downarrow$& PSNR$\uparrow$        & SSIM$\uparrow$      & LPIPS$\downarrow$\\
        \midrule
Bulat \textit{et al.}~\cite{DegradationGAN} & \multirow{4}{*}{4$\times$}& 20.96 & 0.633 & 0.112 & 21.60 & 0.684  & 0.117 &19.45 & 0.616 & 0.222 & 19.63 & 0.592 & 0.228 \\
FSSR~\cite{FSSR} & & 20.46 & 0.680 & 0.102 & 21.63 & 0.683 & 0.095 & 21.40 & 0.653 & 0.208 & 23.32 & 0.649 & 0.136\\
Wei~\textit{et al.}~\cite{DASR}  & & 20.41 & 0.681 & 0.102 & 19.61 & 0.601 & 0.093 & 21.41  & 0.655 & 0.207 & 20.34 & 0.641 & 0.135    \\ 
\midrule
Ours&  & \textbf{21.70}& \textbf{0.722} & \textbf{0.059}& \textbf{23.64}    & \textbf{0.795}    & \textbf{0.093}    & \textbf{22.08} & \textbf{0.675} & \textbf{0.187} & \textbf{23.76}  & \textbf{0.658}  & \textbf{0.127}\\
\bottomrule
\end{tabular}
}
\label{tab:LR}
\vspace{-4pt}
\end{table*}

\begin{table*}[!t]
\caption{SR performance (4$\times$) of our method and other 4 baselines for both handcrafted and real-world degradations.}
\vspace{-8pt}
\centering
\renewcommand\arraystretch{1.0}
\resizebox{0.88\linewidth}{!}{
    \begin{tabular}{l|c|ccc|ccc|ccc|ccc}
        \toprule
        \multirow{3}{*}{Method} & 
        \multirow{3}{*}{Scale} &
        \multicolumn{6}{c|}{Handcrafted degradations}&
        \multicolumn{6}{c}{Real-world degradations}\\
        & &\multicolumn{3}{c|}{Camera sensor noise}  & 
        \multicolumn{3}{c|}{Jpeg compression} &
        \multicolumn{3}{c|}{AIM 2019 dataset}  & 
        \multicolumn{3}{c}{NTIRE 2020 dataset}\\
          & & PSNR$\uparrow$ & SSIM$\uparrow$ & LPIPS$\downarrow$  & PSNR$\uparrow$        & SSIM$\uparrow$      & LPIPS$\downarrow$ & PSNR$\uparrow$        & SSIM$\uparrow$      & LPIPS$\downarrow$& PSNR$\uparrow$        & SSIM$\uparrow$      & LPIPS$\downarrow$\\
        \midrule
Bulat \textit{et al.}~\cite{DegradationGAN} & \multirow{6}{*}{4$\times$}& 17.21 & 0.414 & 0.423 & 16.85 & 0.424  & 0.404 & 19.52 & 0.545 & 0.366 & 21.07 & 0.592 & 0.304\\
FSSR~\cite{FSSR} &  & 20.19 & 0.622 & 0.262 & 21.97 & 0.584 & 0.275 & \textbf{21.84} & \textbf{0.584} & 0.380 & 23.40 & 0.516 & 0.376   \\
Wei~\textit{et al.}~\cite{DASR}  & & 18.85 & 0.593 & 0.261 & 23.25 & 0.644 & 0.271& 21.21  & 0.553 & 0.327 & 20.33 & 0.401 & 0.500 \\ 
Impressionism ~\cite{Impressionism}& &$-$&$-$&$-$&$-$&$-$&$-$& 21.38  & 0.561 & 0.355 & 23.56 & 0.591 & 0.256\\
\midrule
Ours&  & 21.43& 0.668 & 0.207& 23.85 & 0.657 & 0.263 & 21.44 & 0.561 & 0.317 & 23.92  & 0.648  & 0.250 \\
Ours + Sampling  & &\textbf{21.47}&\textbf{0.671}&\textbf{0.201}&\textbf{23.89}&\textbf{0.666}&\textbf{0.258}& 21.46 & 0.565 &\textbf{0.309} & \textbf{23.95} & \textbf{0.659} & \textbf{0.246}\\
\bottomrule
    \end{tabular}}
\label{tab:SR}
\vspace{-5pt}
\end{table*}

\begin{figure*}[!t]
    \centering
    \begin{minipage}[b]{0.81\linewidth}
    \includegraphics[width=1\linewidth]{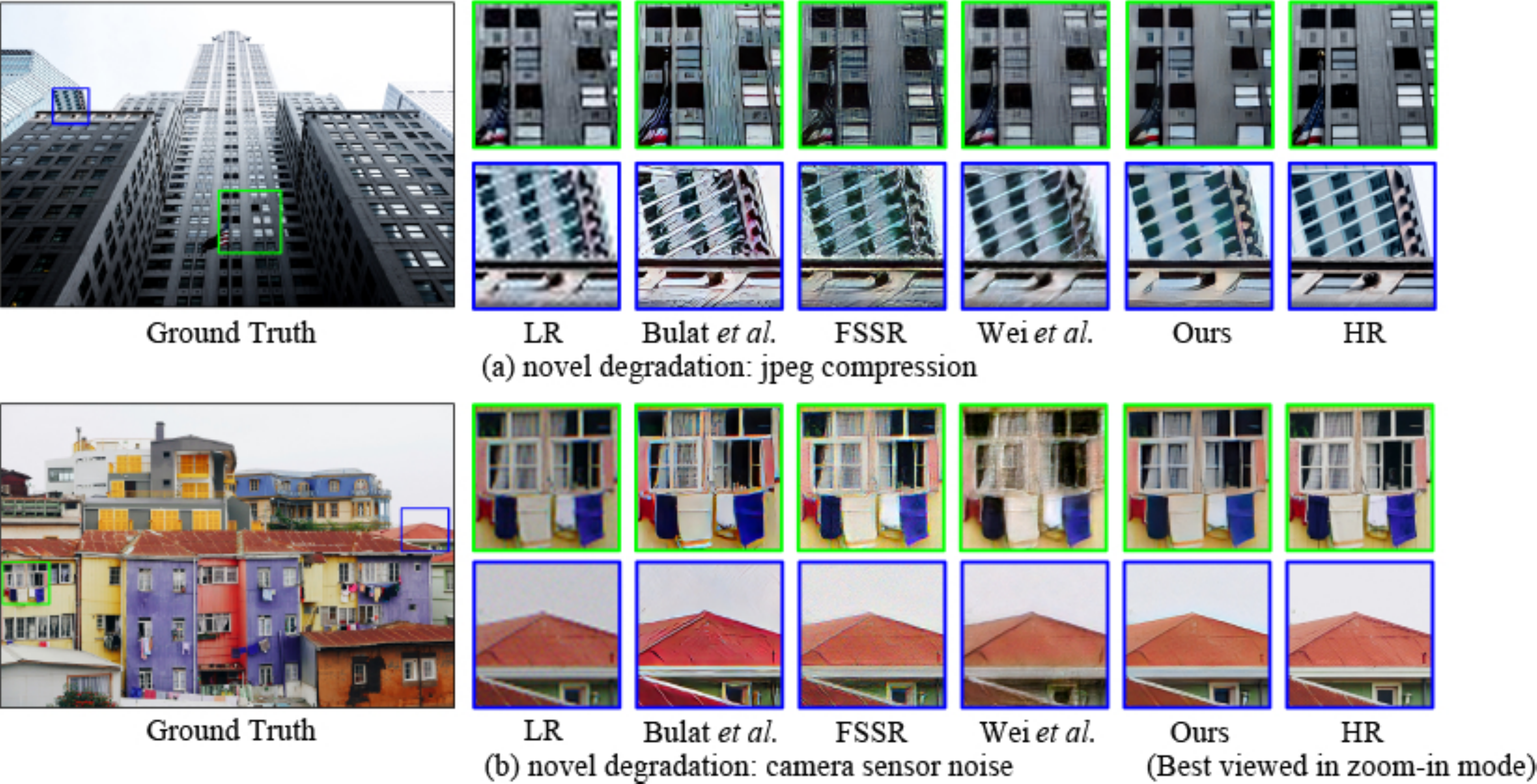}
    \end{minipage}
    \vspace{-8pt}
    \caption{Synthesized SR images (4$\times$) by four methods given LR images with two handcrafted novel degradations, respectively.}
    \vspace{-4pt}
    \label{fig:result2_sy}
\end{figure*}

\begin{figure*}[!t]
    \centering
    \begin{minipage}[b]{0.80\linewidth}
    \includegraphics[width=1\linewidth]{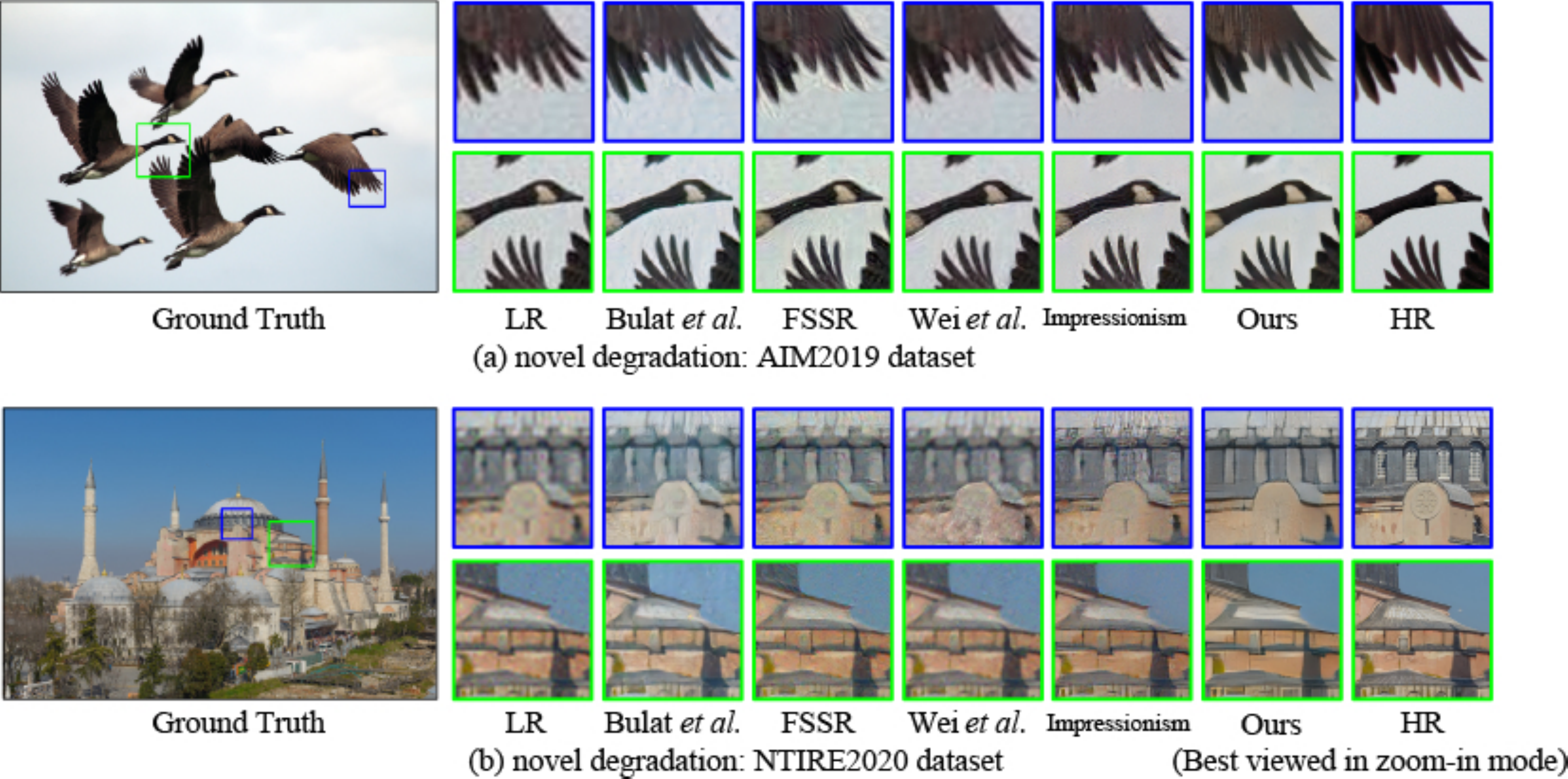}
    \end{minipage}
    \vspace{-8pt}
    \caption{Synthesized SR images (4$\times$) by four methods given LR images with two real-world degradations, respectively.}
    \vspace{-8pt}
    \label{fig:result2_real}
\end{figure*}

\vspace{-8pt}
\subsection{Comparison with State-of-the-art Methods}
Next we compare our method with existing SR methods designed for novel degradations, including 1) Bulat \textit{et al.}~\cite{DegradationGAN} which uses GANs to learn the degradation directly; 2) FSSR~\cite{FSSR} which guarantees the content consistency in the low-frequency domain and ensures the degradation consistency in the high-frequency domain; 3) Wei \textit{et al.}~\cite{DASR} that employs PatchGAN~\cite{PatchGAN} to constrain the generated LR. Impressionism\cite{Impressionism} synthesizes LR images through estimated kernel from real blurry images and noise injection. All these methods follow HR-LR-SR framework to generate HR-LR pairs for training the SR model. For fair comparison, we retrain these methods under our setting for both handcrafted and real-world degradations. 
Both the quality of generated LR and synthesized SR images are evaluated. 

Note that we only present the 4$\times$ scale of SR results in the paper, the 2$\times$ scale of SR results  
are provided in the supplementary materials due to the space limitation.


\noindent\textbf{HR-LR degradation.} It is worth noting that all 3 baseline methods following the HR-LR-SR framework can only deal with one degradation each time by learning one discriminator. In contrast, our method is able to handle various degradations simultaneously leveraging learned degradation representations. 

Table~\ref{tab:LR} presents the performance of generated 4$\times$ scale of degraded LR images by our method and other 3 methods, showing that our method outperforms other methods significantly for both the handcrafted and real-world degradations in terms of all three metrics. These results demonstrate that our method is able to generate more consistent LR with the specified degradation due to the guidance of the degradation representation. We further visualize the generated LR images by different methods for two novel degradations respectively in Figure~\ref{fig:result1_sy} and Figure~\ref{fig:result1_real}. Our method can generate more consistent LR images with the groundtruth LR than other 3 methods, which validates the effectiveness of both learned degradation representations and the HR-LR degradation network.

\noindent\textbf{LR-SR super-resolution.} Table~\ref{tab:SR} presents 4$\times$ scale of SR results of our model and other baselines from degraded LR images for both handcrafted and real-world degradations. Our method outperforms other methods by a large margin in terms of all metrics except `PNSR' and 'SSIM' for `AIM2019 dataset', which demonstrates the superiority of our method over other methods. As analyzed before~\cite{FSSR,DASR}, `LPIPS' measures high-level semantic similarities while `PSNR' and `SSIM' focus on pixel-level similarities, hence `LPIPS' is more consistent with the human perception and more important than `PSNR' and `SSIM' for SR evaluation. Figure~\ref{fig:result2_sy} and Figure~\ref{fig:result2_real} visualize the SR results of different methods for the handcrafted and real-world degradations, respectively. Our method is able to synthesize SR images with much higher quality than other methods. 

\vspace{-8pt}
\section{Conclusion}
In this work we have presented a method for super-resolution of LR images with novel degradations, which cannot be simulated by base degradations with paired training data. Our method first learns a latent representation space for degradations, which can be generalized from base degradations to novel degradations. Then the obtained representations for a novel degradation are leveraged to guide the HR-LR-SR generative process. Extensive experiments validate the effectiveness of the proposed method.
\vspace{-7pt}




\bibliographystyle{ACM-Reference-Format}
\bibliography{sample-base}

\clearpage
\appendix

\section{$2\times$ Scale of Comparative Results with State-of-the-art Methods}

In this section, we present the $2\times$ scale of comparative results between our method and the state-of-the-art methods. Since existing datasets of real-world degradations are mainly constructed on $4\times$ scale, we only conduct experiments on two handcrafted degradations on $2\times$ scales. 

\begin{figure*}[h]
    \centering
    \begin{minipage}[b]{0.95\linewidth}
    \includegraphics[width=1\linewidth]{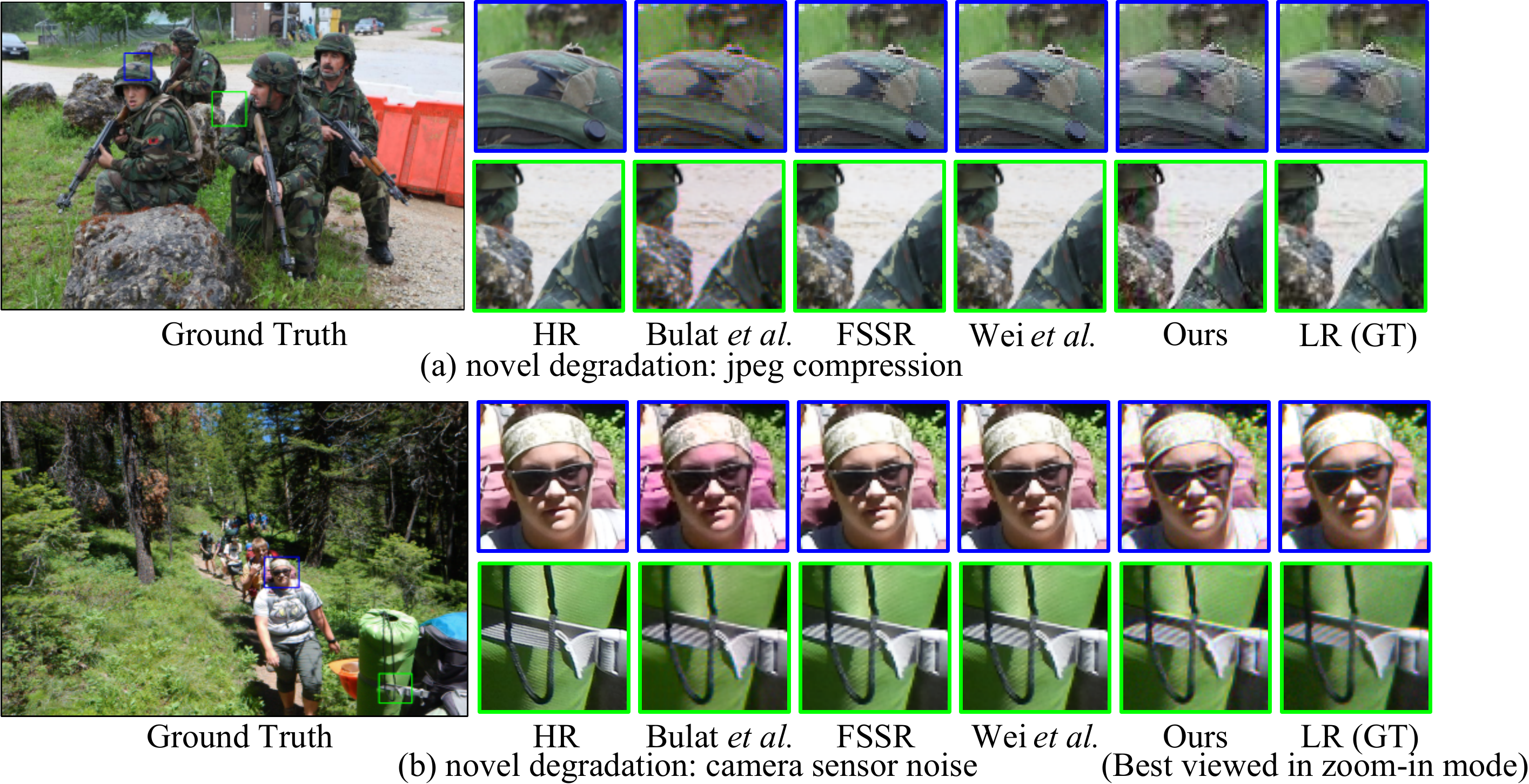}
    \end{minipage}
    \caption{Degraded LR images (2$\times$) by our method and other 3 baselines for two handcrafted novel degradations, respectively.}
    \label{fig:result1_syn_2x}
\end{figure*}

\noindent\textbf{HR-LR degradation.} 
We first visualize the $2\times$ of degraded LR images by our method and other 3 baseline in Figure~\ref{fig:result1_syn_2x}. Our method is able to generate more consistent degraded LR images with the groundtruth LR images than other three methods. 
Table~\ref{tab:LR-x2} presents the quantitative results, which further demonstrate the effectiveness of our method on metrics.

\begin{table}[h]
\caption{Evaluation of degraded LR (2$\times$) by our method and other 3 baselines for both handcrafted degradations.}
\vspace{-8pt}
\centering
\renewcommand\arraystretch{1.0}
\resizebox{1.0\linewidth}{!}{
    \begin{tabular}{l|c|ccc|ccc}
        \toprule
        \multirow{3}{*}{Method} & 
        \multirow{3}{*}{Scale} &
        \multicolumn{6}{c}{Handcrafted degradations} \\
        & &\multicolumn{3}{c|}{Camera sensor noise}  & 
        \multicolumn{3}{c}{Jpeg compression} \\
          & & PSNR$\uparrow$ & SSIM$\uparrow$ & LPIPS$\downarrow$  &  PSNR$\uparrow$        & SSIM$\uparrow$      & LPIPS$\downarrow$\\ \midrule
Bulat \textit{et al.}~\cite{DegradationGAN} & \multirow{4}{*}{2$\times$}& 20.75 & 0.631 & 0.145 & 24.09 & 0.751  & 0.109  \\
FSSR~\cite{FSSR} &  & 21.04 & \textbf{0.681} & 0.140 & 27.25 & 0.8556 & 0.050   \\
Wei~\textit{et al.}~\cite{DASR} &  & 21.28 & 0.6788 & 0.139 & 27.29 & 0.848 & 0.049    \\
\midrule
Ours&  & \textbf{21.92}& 0.643 & \textbf{0.065}& \textbf{29.30} & \textbf{0.892} & \textbf{0.049}  \\
\bottomrule
    \end{tabular}}
\label{tab:LR-x2}
\end{table}

\begin{figure*}[h]
    \centering
    \begin{minipage}[b]{0.95\linewidth}
    \includegraphics[width=1\linewidth]{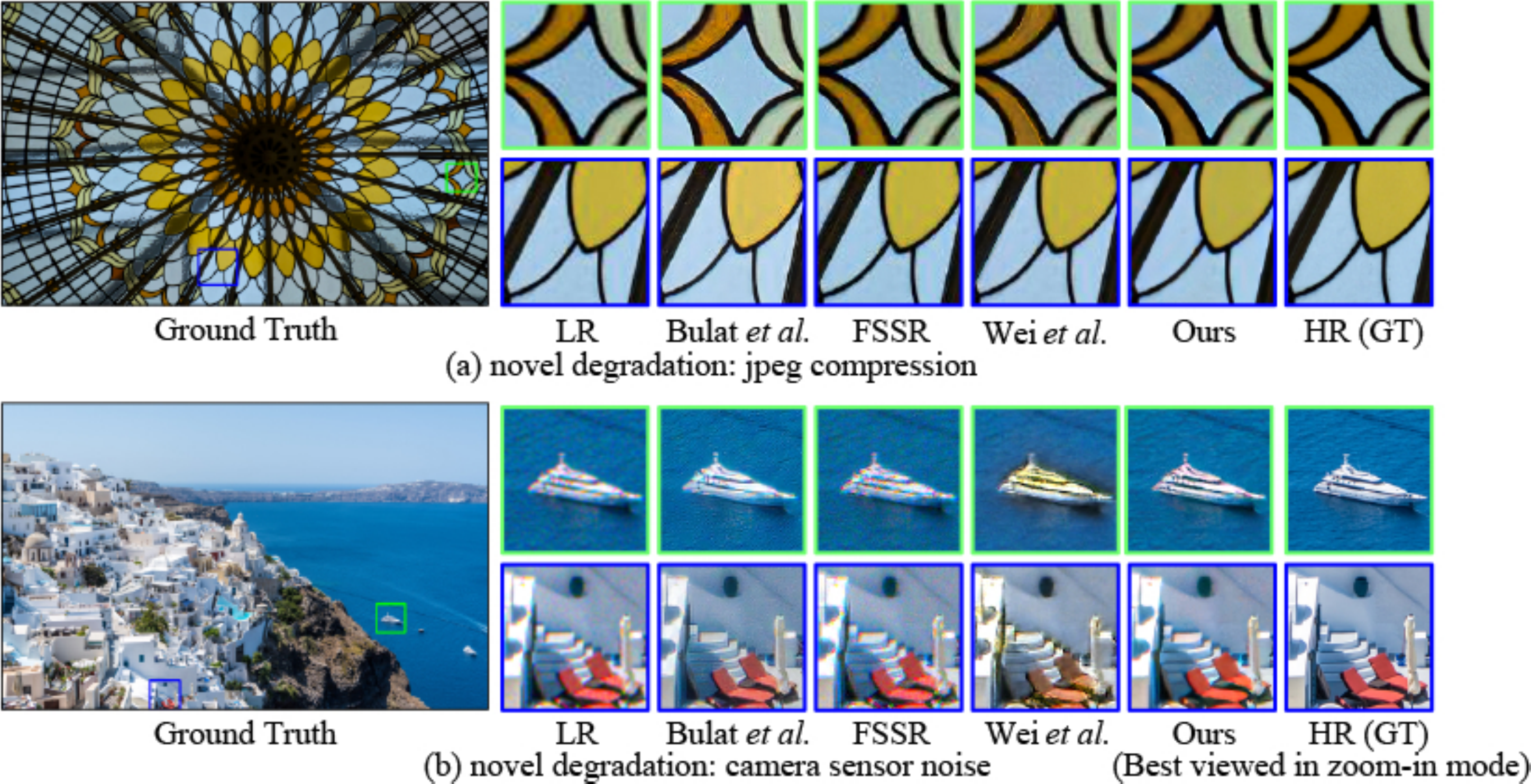}
    \end{minipage}
    \caption{Synthesized SR images (2$\times$) by four methods given LR images with two handcrafted novel degradations, respectively.}
    \label{fig:result2_syn_2x}
\end{figure*}

\noindent\textbf{LR-SR super-resolution.} 
Table~\ref{tab:SR-x2} presents the quantitative performance of $2\times$ SR results by our method and other 3 methods on two degradations. The result shows that our method outperforms other methods, especially on LPIPS, which demonstrates the advantages of our method over other methods. We further visualize the SR images by our method and other baselines. As shown in Figure \ref{fig:result2_syn_2x}, our method can super-resolve images with less compression artifact and less noise on 'jpeg compression' and 'camera-sensor-noise' degradation, respectively. 

\begin{table}[!t]
\caption{SR performance (2$\times$) of our method and other 3 baselines for both handcrafted degradations.}
\vspace{-8pt}
\centering
\renewcommand\arraystretch{1.0}
\resizebox{1.0\linewidth}{!}{
    \begin{tabular}{l|c|ccc|ccc}
        \toprule
        \multirow{3}{*}{Method} & 
        \multirow{3}{*}{Scale} &
        \multicolumn{6}{c}{Handcrafted degradations} \\
        & &\multicolumn{3}{c|}{Camera sensor noise}  & 
        \multicolumn{3}{c}{Jpeg compression} \\
          & & PSNR$\uparrow$ & SSIM$\uparrow$ & LPIPS$\downarrow$  &  PSNR$\uparrow$        & SSIM$\uparrow$      & LPIPS$\downarrow$\\ \midrule
Bulat \textit{et al.}~\cite{DegradationGAN} & \multirow{4}{*}{2$\times$}& \textbf{23.39} & 0.709 & 0.190 & 21.77 & 0.689  & 0.160  \\
FSSR~\cite{FSSR} &  & 20.72 & 0.590 & 0.336 & 26.98 & 0.781 & 0.153   \\
Wei~\textit{et al.}~\cite{DASR} &  & 20.06 & 0.6953 & 0.2349 & 26.53 & 0.783 & 0.160    \\
\midrule
Ours&  & 22.00& \textbf{0.757} & \textbf{0.168}& \textbf{27.09} & \textbf{0.790} & \textbf{0.149}  \\
\bottomrule
    \end{tabular}}
    \vspace{-6pt}
\label{tab:SR-x2}
\end{table}

\section{Ablation Study on Joint Training Scheme for HR-LR Degradation Module}

As explained in Section 3.2 in the paper, our HR-LR degradation module is optimized by a joint training scheme with two individual training modes in parallel. In the first mode, the HR-LR degradation module takes a HR image along with the latent representation of a specified novel degradation as input and generates a LR image which is expected to follow the specified degradation. In the second mode, the HR-LR degradation module acts as a autoencoder, which is supervised to reconstruct an input LR image with the specified novel degradation. Such training mode is conducted to guide the HR-LR degradation module to recognize and reserve the degradation-sensitive features during the encoding and decoding process. 

\begin{figure}[h]
    \centering
    \begin{minipage}[b]{1.0\linewidth}
    \includegraphics[width=1\linewidth]{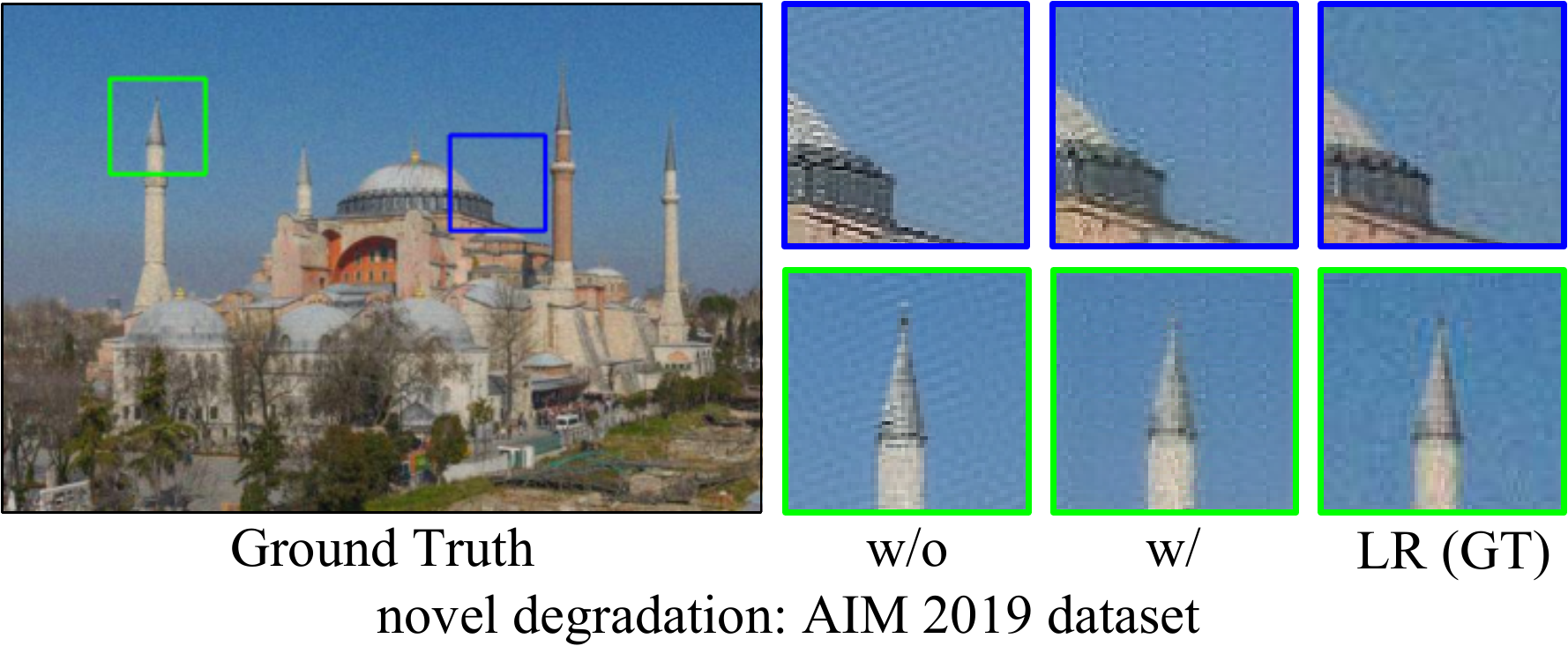}
    \end{minipage}
    \vspace{-10pt}
    \caption{Ablation study on the second training mode of our HR-LR degradation module. The $4\times$ scale of degraded LR images are generated by the HR-LR degradation module of our method, trained with and without the second training mode, respectively.}
    \label{fig:lr_ablation}
\end{figure}

To investigate the effectiveness of such joint training scheme, we visualize the generated $4\times$ scale of degraded LR images on AIM2019 dataset by the HR-LR degradation module of our method in Figure~\ref{fig:lr_ablation}, trained with and without the second training mode, respectively. The result shows that the second training mode leads to more consistent LR results with the groundtruth images.

\end{document}